# Modeling Bilingual Learning Through LLM Training: A Critical-Period Exploration

Uriel Benymon,[1] Tristan Whipps,[1] and Armita Ghobadi[2]

[1]School of General Studies, Columbia University

[2]Barnard College, Columbia University

**Address for correspondence:**

Uriel Benymon

School of General Studies, Columbia University

Department of Computer Science

udb2101@columbia.edu

**Competing interests:** The authors declare none.

**Data availability:** All data supporting the findings of this study are openly available here.

## Abstract

Critical-period accounts of second-language (L2) acquisition suggest that the ability to learn a new language may depend on how easily previously acquired first-language (L1) knowledge can be modified. This exploratory computational study extends a neural language-model approach to critical-period effects by examining internal representations rather than behavior. A small two-layer GPT-2 model was trained on English and then on German (a relatively related L2) or Finnish (a relatively distant L2) through either Sequential training or Elastic Weight Consolidation (EWC), which limits change in parameters important for English. Mean-pooled hidden states for 20 held-out probe sentences per language were compared with those of the English-only model using cosine similarity at the embedding layer and two transformer layers after one and two L2 epochs. English similarity remained high in the transformer layers but became negative in the embedding layer. L2 similarity increased in the transformer layers, and German and Finnish differed more in the second transformer layer than in the first. EWC reduced Fisher-weighted parameter drift by about 6% but changed transformer-layer similarity by less than .001. With one training run per condition, these findings are preliminary.

*Keywords:* bilingualism, critical period, second language acquisition, neural language models, elastic weight consolidation, representational similarity

## Introduction

Bilingual language development has involved continual changes in first-language (L1) and second-language (L2) knowledge, exposure, and use, raising questions about whether similar changes have been represented within the internal structure of neural language models (Constantinescu et al., 2025; Grosjean, 2013). Rather than treating a bilingual as two monolingual speakers combined, Grosjean (2013) characterized bilingual language knowledge as a changing system in which the languages have developed in relation to their history of use. Accordingly, the strengthening and weakening of language knowledge over time has presented bilingual development as a process of shifting organization rather than a fixed endpoint.

Within this changing system, the critical period, defined during the onset of adolescence as the loss of the ability to learn a second language through immersion and the tendency to not forget the first language under deprivation, raised another question concerning how the timing of language exposure and declining plasticity affected L1 retention and L2 acquisition (Constantinescu et al., 2025). Constantinescu et al. (2025) investigated these effects through neural language models, computational systems that learned statistical patterns from language data, by training models on an L1 before introducing an L2 under different learning conditions. Their experiments treated changes in a model's ability to update its parameters as a computational analogue for changes in neuroplasticity, providing controlled access to learning conditions that have been challenging to isolate in human development. Thus, their framework established a route for studying critical-period effects through L1-to-L2 training while retaining control over the timing and flexibility of language learning.

One method for restricting this flexibility emerged through Elastic Weight Consolidation (EWC), which reduces changes to parameters that were important for previously learned information (Kirkpatrick et al., 2017). Kirkpatrick et al. (2017) introduced EWC as an approach for limiting catastrophic forgetting, where neural networks lose earlier knowledge after learning a new task. Subsequently, Constantinescu et al. (2025) applied EWC to language acquisition as a computational analogue of reduced plasticity during later learning. This connection made sequential L1-to-L2 training useful for examining not only whether an L2 was learned, but also how much earlier L1 knowledge remained stable as new language information entered the network.

Still, behavioral performance alone has not fully described what changes inside a model during this process. Internal representations, or patterns of neural activation that are produced as linguistic input moves through a network, offer another level at which language learning could be examined. Liu et al. (2019) found that neural language models encoded different forms of linguistic information across their internal layers, showing that representations did not remain uniform throughout a network. Consequently, tracking multiple layers across L2 training has offered a way to ask whether language-related change has occurred evenly or followed distinct representational trajectories at different depths of the model.

Beyond locating those representations, prior work also established methods for comparing how neural representations differ between models or stages of learning. Kornblith et al. (2019) evaluated several quantitative approaches to representational similarity, including the Centered Kernel Alignment (CKA), and argued that simpler similarity measures can be misleading if applied without care. The present study adopts cosine similarity rather than CKA because it directly captures the orientation of individual sentence representations relative to a pre-L2 baseline, which suits a longitudinal, per-sentence comparison. This choice carries a cost that Kornblith et al.'s (2019) critique makes explicit: cosine similarity cannot distinguish genuine linguistic reorganization from general shifts in the geometry of the representational space, a limitation addressed directly in the discussion.

Language learning has also depended on the relationship between the languages involved, making the selection of L1 and L2 conditions important for interpreting representational change (Grosjean, 2013). Littell et al. (2017) developed URIEL and lang2vec[1] to represent languages through features including typological, geographical, and phylogenetic information, allowing cross-language relationships to be expressed computationally. In this framework, typological information provided a principled basis for comparing languages instead of labeling pairs as related or unrelated through intuition alone. Accordingly, such language features offered a bridge between Grosjean's (2013) account of interacting bilingual language systems and a measurable comparison of how different L2 conditions affected a shared L1-trained network.

Despite these developments, prior research left an important intersection less resolved: how internal representations change throughout sequential L2 acquisition, whether those changes differ across layers, and whether language relatedness corresponds with different trajectories. Constantinescu et al. (2025) supplied the critical-period training framework, while Liu et al. (2019) and Kornblith et al. (2019) showed that internal representations contained measurable structure that could be compared. Likewise, Littell et al. (2017) supplied a way to ground cross-linguistic relationships in linguistic features. Taken together, these strands motivated an extension from asking whether a neural language model retained or acquired a language toward examining where and how its internal representations changed during that learning process.

The present study examines this gap through an English L1 model that learns both German and Finnish, separately, as an L2 under Sequential and EWC-constrained training conditions. German and Finnish were selected to mirror Constantinescu et al. (2025), reflecting the same criteria: German's shared language family with English, Finnish's lack of relation to either, and all three languages' high-resource status. Internal change is measured at the embedding layer and two transformer layers across an early L2 checkpoint and the final training stage, while cosine similarity to the English-only (pre-L2) model tracks change in the representations of English probe sentences, referred to here as L1 representational retention, and of German or Finnish probe sentences; neither is a direct measure of language proficiency. As seen in Table 1, German and Finnish provide contrasting L2 conditions through which differences in representational trajectories can be considered alongside cross-linguistic relatedness (Littell et al., 2017). In this way, the experiment connects bilingual development, constrained plasticity, language relatedness, and internal neural change within one controlled computational design.

---

[1]A database providing information-rich vector identifications of languages drawn from typological, geographical, and phylogenetic databases, primarily representing binary language facts (Littell et al., 2017).

The study addresses three research questions. The first is how do the internal representations of an English-trained neural language model change across early and final stages of German or Finnish L2 acquisition? Secondly, how does representational change differ between German and Finnish conditions across the embedding layer and transformer layers? And finally, to what extent does EWC-constrained L2 training preserve English L1 representations differently from unconstrained Sequential training? We hypothesize that the internal representations will change across L2 training, but the magnitude and direction of change will differ by representational layer and L2 condition. More specifically, German and Finnish will produce different trajectories of representational similarity, consistent with the expectation that cross-linguistic differences will shape how new language information becomes organized within the model (Littell et al., 2017). Finally, EWC-constrained models will retain greater similarity to their English L1 representations than sequentially trained models because EWC will restrict changes to parameters important for earlier learning (Constantinescu et al., 2025; Kirkpatrick et al., 2017). These predictions are directional and falsifiable: equivalent trajectories across languages, layers, or training conditions will provide evidence against the corresponding hypotheses. The parameter-level analysis of EWC reported in Section 3.3 was conducted after these hypotheses were specified and is presented as exploratory.

# Methods

## 2.1 Study Design and Experimental Overview

This study used a computational experiment to investigate how the internal representations of a language model changed during L2 acquisition and whether these changes differed across language and training conditions. The design extended the neural language-model framework of Constantinescu et al. (2025), which modeled critical-period effects by manipulating when a model encountered another language and how freely its parameters changed during later training. The present study did not vary when the L2 was introduced; it held the L1-to-L2 schedule fixed and examined internal representational change within this framework, so it does not itself test for a critical-period boundary. English served as the L1, while German and Finnish formed two L2 conditions intended to provide different relationships with English. Accordingly, the experiment compared changes in both English representational retention and L2 representations across ordinary Sequential training and training constrained through Elastic Weight Consolidation (EWC).

## 2.2 Model and Training Framework

The experiment used the GPT-2 language-model architecture and training framework adapted from Constantinescu et al. (2025). The model first learned English before continuing through an L2 training stage with either German or Finnish data, allowing the English-only checkpoint to function as the pre-L2 baseline. Representations were examined at Epoch 1 and Epoch 2 to capture an early and later point in the acquisition trajectory. This longitudinal comparison provided a controlled way to examine whether representations remained stable or shifted as exposure to the L2 continued. The model was a small GPT-2 architecture with two transformer layers, two attention heads, and 128-dimensional hidden states (6,960,768 parameters, with tied input and output embeddings), paired with a custom byte-level BPE tokenizer of 8,016 tokens; because the model configuration retained the standard GPT-2 vocabulary size of 50,257, embedding rows beyond the tokenizer's vocabulary were never used. The English L1 model was trained from scratch, and each of the four L2 runs (German or Finnish, Sequential or EWC) was initialized from that same saved English checkpoint, as recorded in each L2 checkpoint's configuration metadata. All runs used seed 42, two epochs, a learning rate of 0.001 with a warmup ratio of 0.07, a per-device batch size of 4 without gradient accumulation, and evaluation and checkpoint saving at the end of each epoch. Because the corpora were matched by sentence count rather than token count, the number of completed optimization steps differed by language: 70 for English L1, 170 for each German L2 run, and 138 for each Finnish L2 run. Epoch 1 representations were taken from the checkpoint saved at the end of the first L2 epoch (e.g., step 85 for German), and Epoch 2 (Final) representations from the model saved at the end of training; the best-validation checkpoint was not substituted. Checkpoint metadata records Transformers version 4.30.0.dev0, and the evaluation reported here was run with Python 3.10.20, PyTorch 2.0.1, and Accelerate 0.19.0.

**Table 1**

*URIEL/lang2vec Measures Used to Select the L2 Language Conditions*

| L2 language | Syntactic distance from English | Shared genealogical features with English | Experimental classification |
|---|---|---|---|
| German | 0.0777 | 4 | Relatively related L2 |
| Finnish | 0.2136 | 0 | Relatively distant L2 |

*Note.* Syntactic distance and genealogical features were obtained using URIEL/lang2vec (Littell et al., 2017). Syntactic distance is the proportion of the 103 URIEL syntactic features, as imputed by lang2vec's k-nearest-neighbor method, on which the language differs from English (8 of 103 for German and 22 of 103 for Finnish); greater values indicate greater syntactic difference from English. This proportion is not lang2vec's precomputed syntactic distance (.42 for German and .53 for Finnish), although both give the same ordering. Shared genealogical features indicate the number of URIEL language-family features shared with English (for German: Indo-European, Germanic, Northwest Germanic, and West Germanic). German was classified as the relatively related L2 condition, whereas Finnish was classified as the relatively distant L2 condition.

## 2.3 Language Conditions and Typological Comparison

German and Finnish served as contrasting L2 conditions based on syntactic and genealogical measures obtained through URIEL/lang2vec (Littell et al., 2017). German showed a syntactic distance of 0.0777 from English and shared four genealogical features with English, whereas Finnish showed a larger syntactic distance of 0.2136 and shared no genealogical features with English (see Table 1). Because typological distance is not the only systematic difference between the two L2 conditions, probe-sentence length was also examined using the custom byte-level BPE tokenizer employed by the model, rather than the original GPT-2 tokenizer. Contrary to the expectation that Finnish's agglutinative morphology would produce substantially longer token sequences, the German and Finnish probe sets were nearly matched in average tokenization length. German averaged 59.50 tokens per sentence (SD = 38.98) and Finnish averaged 57.50 tokens per sentence (SD = 24.07), each based on the 20 probe sentences per language described in Section 2.5. This makes a simple difference in average sequence an unlikely explanation for any German-Finnish contrast in the results, though it does not establish that the two languages were segmented equivalently at the subword level. Finnish is also lower-resource than English or German in most large-scale training corpora; the training corpora here were matched by sentence count rather than by verified token count (Section 2.5), so language relatedness and training-data exposure cannot be fully separated in the current design.

**2.4 Sequential and EWC Training Conditions**

Each language condition contained Sequential and EWC training variants. Sequential training allowed the model to continue from English training into the L2 without an added mechanism for protecting parameters associated with prior knowledge. Conversely, EWC penalized changes to parameters according to their estimated importance for previously acquired information, a method originally developed to reduce catastrophic forgetting in neural networks (Kirkpatrick et al., 2017). Constantinescu et al. (2025) used this constraint as a computational manipulation of reduced plasticity during later language learning. Comparing these conditions therefore allowed us to examine whether restricting parameter change preserved L1 representations or altered the development of L2 representations. In the training code adapted from Constantinescu et al. (2025), the EWC condition added a quadratic penalty to the language-modeling loss, $L_{\mathrm{total}} = L_{\mathrm{LM}} + \lambda \, \Sigma_i \, F_i(\theta_i - \theta^*_i)^2$, where $\theta^*$ denotes the parameters of the English L1 model, $F_i$ is the estimated Fisher information for parameter $i$, and $\lambda = 20$, the default value in the code, which provides no command-line override. This penalty omits the conventional factor of one half and is applied only in training mode, so evaluation losses computed during training do not include it. The diagonal Fisher information was estimated once after English L1 training on the English validation data by sampling target tokens from the model's own predictive distribution (10 samples per batch) and accumulating squared gradients; the accumulated values were divided by the number of samples but not by the number of batches, so their absolute scale depends on the number of validation batches processed. The saved EWC checkpoints use the penalized model class, and the Fisher matrix saved with the English checkpoint was structurally compatible with the model, but the magnitude of the penalty was not logged during training.

### 2.5 Probe Materials and Hidden-State Extraction

Training and probe materials were drawn from the Helsinki-NLP OPUS Books parallel corpora (Tiedemann, 2012). Each language's corpus contained 3,000 cleaned, unique sentences divided into 2,490 training, 255 validation, and 255 test sentences. English and German sentences were sampled from disjoint rows of the German-English parallel corpus, while Finnish sentences were sampled independently from the English-Finnish parallel corpus, so the three probe sets were not parallel translations of one another. The corpora were drawn from the upstream Hugging Face release of OPUS Books without pinning a specific dataset revision, so the fixed sampling seed alone does not guarantee identical corpora if the upstream data change. The shared tokenizer was trained on text that included the held-out validation and test splits; this is tokenizer-level exposure to evaluation text rather than gradient-based training on test examples, but it means the tokenizer was not fully independent of the probe sentences. From each held-out test set, 20 probe sentences per language were reproducibly sampled using a fixed random seed (42), yielding the English, German, and Finnish probe sets used throughout the representational analysis. Probe sentence length, measured with this same tokenizer, averaged 19.80 tokens per sentence for English (SD = 16.56), 59.50 for German (SD = 38.98), and 57.50 for Finnish (SD = 24.07). English probes measure changes to L1 representations after L2 exposure, while the German and Finnish probes tracked representations of the language being acquired. For every sentence, hidden states were extracted from the embedding layer, transformer layer 1, and transformer layer 2 at each selected checkpoint. Attention-mask-aware mean pooling combined token-level activations into one sentence representation for each layer, while evaluation mode and disabled gradient computation prevented the extraction procedure from changing the model. Probe sentences were tokenized with a maximum length of 256 tokens, which accommodated every probe; the longest, a German sentence, contained 177 tokens. An earlier extraction used a 128-token limit that truncated this sentence, so all German representations were re-extracted from the same saved checkpoints at full length; the vectors for the other 19 German sentences were unchanged, and all results reported here use the corrected extraction. Hidden states provide an intermediate view of how neural networks encode information rather than relying exclusively on their final behavioral output, making comparisons of these representations useful for examining model change (Liu et al., 2019).

**2.6 Representational Similarity Measures**

Cosine similarity was used to quantify representational change, which compared the orientation of corresponding sentence vectors between a training checkpoint and its pre-L2 baseline. For a probe sentence $s$ and layer $\ell$, similarity was computed as $\cos(h_t, h_0) = (h_t \cdot h_0) / (\|h_t\| \|h_0\|)$, where $h_t$ is the mean-pooled representation of $s$ at checkpoint $t$ and $h_0$ is the representation of the same sentence produced by the English-only model. Values range from −1 to 1 and can be negative, as occurred in the English embedding layer (Table 2). For L1 retention,

English representations at Epoch 1 and Epoch 2 were compared with representations from the English-only model; higher similarity therefore indicated that an English representation remained closer in orientation to its original English-only representation. For L2 analysis, German and Finnish representations at the same stages were compared with their respective representations produced by the pre-L2 English-only model. Consequently, lower similarity indicated greater departure from the baseline representation, while changes from Epoch 1 to Epoch 2 revealed the direction of the L2 trajectory.

**2.7 Data Analysis**

Sentence-level cosine similarities were aggregated by language condition, training condition, stage, and layer before the primary contrasts were examined. First, Epoch 1-to-Epoch 2 changes measured representational development across L2 training. Next, EWC − Sequential contrasts tested whether the plasticity manipulation changed either English retention or L2 representations, while Finnish − German contrasts examined differences associated with the two language conditions. Ninety-five percent confidence intervals were obtained by percentile bootstrap with 10,000 resamples (seed 42), with the probe sentence as the resampling unit. Stage contrasts, EWC − Sequential contrasts, and Finnish − German contrasts on English probes compared the same sentences across models, so sentence-level differences were resampled and effect sizes are reported as paired Cohen's $d_z$. Because the German and Finnish probe sets contained different, non-parallel sentences, Finnish − German contrasts on L2 probes treated the two languages as independent groups, resampled each separately, and report Hedges' $g$. These statistics describe variation among the probe sentences within the present trained model run; therefore, they do not represent uncertainty across independently trained model seeds and remain exploratory measures of the observed representational patterns. An interval that excludes zero therefore indicates a difference that was consistent across these 20 probes for this run, not a difference expected to replicate across independently trained models. No adjustment was made for multiple comparisons across the 72 contrasts computed (spanning layers, stages, training conditions, and contrast types), so intervals that exclude zero should be read as descriptive rather than confirmatory.

**Table 2**

*Mean Cosine Similarity of L1 and L2 Representations Across L2 Training Conditions*

| Analysis | L2 | Training condition | Stage | Embedding | Transformer layer 1 | Transformer layer 2 |
|---|---|---|---|---|---|---|
| **English L1 retention** | German | Sequential | Epoch 1 | -0.0174 | 0.7721 | 0.8624 |
| | German | Sequential | Epoch 2 | -0.1339 | 0.7633 | 0.8609 |
| | German | EWC | Epoch 1 | -0.0080 | 0.7726 | 0.8625 |
| | German | EWC | Epoch 2 | -0.1207 | 0.7638 | 0.8616 |
| | Finnish | Sequential | Epoch 1 | 0.0037 | 0.8058 | 0.8593 |
| | Finnish | Sequential | Epoch 2 | -0.1091 | 0.7945 | 0.8492 |
| | Finnish | EWC | Epoch 1 | 0.0096 | 0.8067 | 0.8587 |
| | Finnish | EWC | Epoch 2 | -0.0982 | 0.7945 | 0.8490 |
| **L2 representation** | German | Sequential | Epoch 1 | 0.6780 | 0.7915 | 0.8765 |
| | German | Sequential | Epoch 2 | 0.6674 | 0.8497 | 0.9148 |
| | German | EWC | Epoch 1 | 0.6781 | 0.7911 | 0.8761 |
| | German | EWC | Epoch 2 | 0.6675 | 0.8500 | 0.9147 |

| Analysis | L2 | Training condition | Stage | Embedding | Transformer layer 1 | Transformer layer 2 |
|---|---|---|---|---|---|---|
| | Finnish | Sequential | Epoch 1 | 0.5149 | 0.7800 | 0.7039 |
| | Finnish | Sequential | Epoch 2 | 0.5071 | 0.8162 | 0.8103 |
| | Finnish | EWC | Epoch 1 | 0.5150 | 0.7800 | 0.7038 |
| | Finnish | EWC | Epoch 2 | 0.5072 | 0.8162 | 0.8099 |

*Note.* Values represent mean cosine similarity across 20 probe sentences. For English L1 retention, post-L2 English representations were compared with their corresponding representations at the English L1 baseline. For L2 representation, German and Finnish representations after L2 training were compared with representations produced for the same probe sentences by the English-only model before L2 training. Higher values indicate greater representational similarity to the corresponding pre-L2 baseline. Sequential refers to ordinary L2 training; EWC = elastic weight consolidation. Epoch 2 represents the final L2 checkpoint. The embedding output and both transformer layers consisted of 128-dimensional hidden-state representations.

# Results

## 3.1 English L1 Retention

In the transformer layers, English representations retained high similarity to the pre-L2 baseline throughout L2 training in both language conditions (between .76 and .87; see Table 2), whereas embedding-layer similarity was already near zero at Epoch 1 and became negative by Epoch 2 (-.1339 for German and -.1091 for Finnish under Sequential training). Mean similarity declined from Epoch 1 to Epoch 2 in every layer and condition, with the largest decline occurring in the embedding layer. Under Sequential training at Epoch 2, Finnish showed higher

English-retention similarity than German in transformer layer 1, with a mean difference of .0311, 95% CI [.0153, .0478], whereas the confidence interval for transformer layer 2 included zero, with a mean difference of -.0118, 95% CI [-.0240, .0025]. These layer-specific patterns indicate that English representational similarity did not shift uniformly across the network.

### 3.2 L2 Representational Change

German and Finnish representations both became more similar to their pre-L2 baseline between Epoch 1 and Epoch 2 within the transformer layers (see Table 2), whereas embedding-layer similarity decreased slightly in both languages (by about .011 for German and .008 for Finnish). Because higher similarity to this baseline indicates that a representation more closely resembles what an English-only model would have produced for the same sentence, this pattern indicates that, in cosine orientation, transformer-layer L2 representations moved toward, rather than away from, the pre-L2 baseline as training progressed; it does not by itself indicate improved L2 acquisition. The largest such change occurred for Finnish in transformer layer 2, where similarity increased by approximately .11 across training (from .7039 to .8103 under Sequential training; mean change = .1064, 95% CI [.0946, .1176]).

### 3.3 Training Condition (EWC vs. Sequential)

EWC and Sequential training produced nearly overlapping trajectories in the transformer layers for both English L1 retention and L2 representations (see Table 2): across languages and stages, the largest absolute EWC − Sequential difference in transformer-layer mean similarity was less than .001. These differences were small but not absent; the sentence-level 95% CI for the paired EWC − Sequential difference excluded zero in seven of eight transformer-layer comparisons for English probes and in six of eight for L2 probes. The embedding layer showed larger differences for English probes, where EWC retained higher similarity to the English baseline than Sequential training at both stages, most clearly at Epoch 2 (by .0132 for German and .0110 for Finnish). Thus, hidden-state differences between the training conditions were small rather than absent, and were largest in the English embedding layer.

A separate parameter-level analysis, conducted after the original analysis, compared the saved final L2 parameters with the English L1 parameters, weighting each squared parameter change by its estimated Fisher information. EWC reduced this Fisher-weighted drift relative to Sequential training by about 6% for both German (0.000328 vs. 0.000350) and Finnish (0.000349 vs. 0.000371), whereas unweighted Euclidean distance from the English parameters differed by less than 1% (German: 141.38 vs. 142.18; Finnish: 146.41 vs. 146.99). This pattern is consistent with the EWC penalty constraining the parameters it was designed to protect, but it is a parameter-level measure that is distinct from hidden-state similarity: it does not show that English proficiency was better retained, and it does not establish that the penalty was active at every optimization step.

**3.4 Language Condition (German vs. Finnish)**

German and Finnish diverged most clearly within the deeper transformer layers at Epoch 2. Under Sequential training, the Finnish − German difference was small, and its own confidence interval crossed zero in transformer layer 1 with a mean difference of -0.0335, 95% CI [-0.0729, 0.0037], Hedges' $g$ = -0.5181, but was substantially larger and excluded zero in transformer layer 2 with a mean difference of -0.1045, 95% CI [-0.1186, -0.0909], Hedges' $g$ = -4.4256. Because the German and Finnish probes are different sentences, these are between-group comparisons that can reflect differences in sentence content and tokenization as well as language condition, and the large standardized effect in transformer layer 2 reflects the low variability of layer-2 similarity across probes as much as the absolute difference of about .10; it should not be read as a large behavioral effect. Taken together with Table 2, L2 representations became more like their pre-L2 baseline over training, and German and Finnish trajectories diverged most clearly in the deeper transformer layer.

## 4. Discussion

The first research question asked how the internal representations of an English-trained model change across early and final stages of German or Finnish L2 acquisition. English L1 representations remained comparatively stable across the transformer layers, though not in the embedding layer, while representations of the L2 being acquired became more like their pre-L2 baseline over the same period (Table 2). In terms of cosine orientation, this indicates that German and Finnish sentence representations came to resemble what an English-only model would have produced *more* closely as training continued, rather than diverging from it. One possible but untested explanation is a general property of training rather than a language-specific learning signal: if hidden-state representations became more concentrated in shared directions as training progressed, cosine similarity to the baseline could rise regardless of the linguistic content being processed. Representational dispersion was not measured in the present study, and because cosine similarity captures representational alignment but cannot by itself explain what linguistic information a hidden state encodes (Kornblith et al., 2019), the present design cannot distinguish this account from L2-specific change since it did not include a control comparison. A control comparison such as unrelated or scrambled sentences would reveal whether the same increase occurs independent of what the model is actually learning about German or Finnish. This computational view of representational change nonetheless complements the behavioral focus of prior critical-period modeling (Constantinescu et al., 2025), which examined L1-to-L2 training effects primarily through model performance rather than internal representational structure.

The second research question asked how representational change differed between German and Finnish across the embedding and transformer layers. The two conditions diverged most clearly in the deeper layers: Finnish showed the larger increase in transformer layer 2, while German showed the larger increase in transformer layer 1. The Finnish − German difference was small, and its confidence interval crossed zero in transformer layer 1, with a mean difference of -0.0335, 95% CI [-0.0729, 0.0037], Hedges' *g* = -0.5181, but was substantially larger and excluded zero in transformer layer 2, with a mean difference of -0.1045, 95% CI [-0.1186, -0.0909], Hedges' *g* = -4.4256. This divergence is compatible with language relatedness shaping how new material is integrated at different depths of the network, since German and Finnish were selected to represent contrasting typological distances from English using the URIEL/lang2vec framework (Littell et al., 2017). However, German and Finnish differ in more than typological distance. Average tokenized sentence length was similar between the two probe sets (59.50 and 57.50 tokens), but similar averages do not rule out tokenization-based explanations, because the length distributions, subword segmentation, and lexical content of the two non-parallel probe sets differ. Training exposure also differed: the German runs completed 170 optimization steps and the Finnish runs 138. Finnish also remains lower-resource than German in typical training corpora, and neither the fine-grained subword composition of the two languages nor the total token count of the full training corpora has been independently verified, so language relatedness cannot yet be isolated as the sole cause of the layer-2 difference. This qualification is consistent with Grosjean's (2013) broader caution against treating a bilingual system's two languages as isolated, cleanly separable conditions rather than a jointly developing whole.

The third research question asked to what extent the EWC-constrained training preserved English L1 representations differently from unconstrained Sequential training. EWC and Sequential training produced nearly identical trajectories in the transformer layers, where the largest difference in mean similarity was under .001, whereas the English embedding layer showed somewhat larger differences, with EWC retaining higher similarity to the English baseline (Table 2). EWC restricts changes to parameters estimated to be important for previously learned information, a mechanism originally developed to reduce catastrophic forgetting in neural networks (Kirkpatrick et al., 2017), and Constantinescu et al. (2025) applied this same constraint as a computational manipulation of reduced plasticity during L2 learning. The small differences observed here suggest that, under the present training regime, EWC's parameter-level constraint translated into only slight differences in the transformer-layer hidden-state representations examined, with somewhat larger differences at the English embedding layer. At the parameter level, however, EWC reduced Fisher-weighted drift from the English parameters by about 6% in both languages (Section 3.3), so the constraint did alter the parameters it targeted; this parameter-level difference did not translate into a large change in the sentence-level cosine measure used here. A relatively weak penalty in this small model, given $\lambda = 20$ and the Fisher scaling described in Section 2.4, is one possible reason. The constraint may also be concentrated in parameters or layers that this probe-based analysis, which drew on the layer-comparison approach of Liu et al. (2019), was not positioned to capture.

These findings remain exploratory: the analysis is based on a single trained-model seed, a single small two-layer architecture, and 20 probe sentences per language, and cosine similarity alone cannot distinguish genuine linguistic reorganization from general drift in the representational space (Kornblith et al., 2019). In addition, cosine similarity to the English-only baseline is not a behavioral measure of L1 retention or L2 proficiency, and the design did not vary the age or timing of L2 onset, so the results do not bear directly on human critical-period effects. Further limitations include tokenizer training on text that included the held-out splits, an unpinned upstream dataset revision, corpora matched by sentence count rather than token count, and the absence of a direct behavioral measure such as held-out loss or perplexity. Future work extending the design to multiple seeds, additional language pairs, and a scrambled-sentence control would help determine whether the patterns reported here reflect L2-specific learning or broader training dynamics.

**Research Status and Limitations**

The data consist of a single training run (seed 42) for each of five model conditions (English L1, and German and Finnish L2 under Sequential and EWC training), training corpora of 2,490 sentences per language, and 20 probe sentences per language (n = 60 probe sentences in total). The study was conducted as part of a course-based research project at Barnard College, Columbia University.[2] Confidence intervals describe variation among these probe sentences within a single run, not across independently trained models, and were not corrected for multiple comparisons. The model is a small two-layer architecture, the German and Finnish probes are not parallel translations, training exposure differed by language, and cosine similarity is not a behavioral measure of L1 retention or L2 proficiency. The study did not vary the timing of L2 onset and therefore does not test human critical-period effects. These results should be treated as a basis for further study rather than unequivocal findings.

[2]The project was developed as part of *Bilingual Minds: Language, Brain & Culture Across Development*, a four-credit seminar offered at Barnard College, Columbia University. Additional course information is available in the Columbia University course directory.